\documentclass{article}
\usepackage[final]{neurips_2024}
\usepackage[utf8]{inputenc} 
\usepackage[T1]{fontenc}    
\usepackage[hidelinks]{hyperref}      
\usepackage{url}            
\usepackage{booktabs}       
\usepackage{amsfonts}       
\usepackage{mathtools}
\usepackage{nicefrac}      
\usepackage{microtype}     
\usepackage{xcolor}         
\usepackage{natbib}

\usepackage{graphicx}       
\usepackage{xcolor}
\usepackage{soul}
\usepackage{subcaption} 
\usepackage{soul}

\title{Dreaming Learning}

\author{
  Alessandro Londei\\
 Sony Computer Science Laboratories - Rome\\
 Joint Initiative CREF-SONY, Centro Ricerche Enrico Fermi\\
Via Panisperna 89/A, 00184, Rome, Italy \\
  \texttt{alessandro.londei@sony.com} \\
\And
Matteo Benati \\
 Department of Computer, Automatic and Management Engineering \\
  Sapienza University, Via Ariosto 25, Rome, Italy \\
 \texttt{matteo.benati@uniroma1.it} \\
 \And
Denise Lanzieri \\
 Sony Computer Science Laboratories - Rome\\
 Joint Initiative CREF-SONY, Centro Ricerche Enrico Fermi\\
Via Panisperna 89/A, 00184, Rome, Italy \\
  \texttt{denise.lanzieri@sony.com} \\
\And
Vittorio Loreto \\
 Sony Computer Science Laboratories - Rome\\
 Joint Initiative CREF-SONY, Centro Ricerche Enrico Fermi\\
 Via Panisperna 89/A, 00184, Rome, Italy, \\
 Centro Ricerche Enrico Fermi (CREF) \\
Via Panisperna 89/A, 00184, Rome, Italy ,\\
Sapienza University of Rome, Physics Department\\
Piazzale A. Moro, 2, 00185, Rome, Italy, \\
Complexity Science Hub \\
Josefstädter Strasse 39, A 1080, Vienna, Austria \\
  \texttt{vittorio.loreto@sony.com} \\
}

\begin{document}
\maketitle
\begin{abstract}
    Incorporating novelties into deep learning systems remains a challenging problem. Introducing new information to a machine learning system can interfere with previously stored data and potentially alter the global model paradigm, especially when dealing with non-stationary sources. In such cases, traditional approaches based on validation error minimization offer limited advantages. To address this, we propose a training algorithm inspired by Stuart Kauffman’s notion of the Adjacent Possible. This novel training methodology explores new data spaces during the learning phase. It predisposes the neural network to smoothly accept and integrate data sequences with different statistical characteristics than expected. The maximum distance compatible with such inclusion depends on a specific parameter: the sampling temperature used in the explorative phase of the present method. This algorithm, called Dreaming Learning, anticipates potential regime shifts over time, enhancing the neural network’s responsiveness to non-stationary events that alter statistical properties. To assess the advantages of this approach, we apply this methodology to unexpected statistical changes in Markov chains and non-stationary dynamics in textual sequences. We demonstrated its ability to improve the auto-correlation of generated textual sequences by $\sim$  29\% and enhance the velocity of loss convergence by $\sim$ 100\% in the case of a paradigm shift in Markov chains.
\end{abstract}
\section{Introduction}
\begin{figure}
    \centering
    \includegraphics[width=\columnwidth]{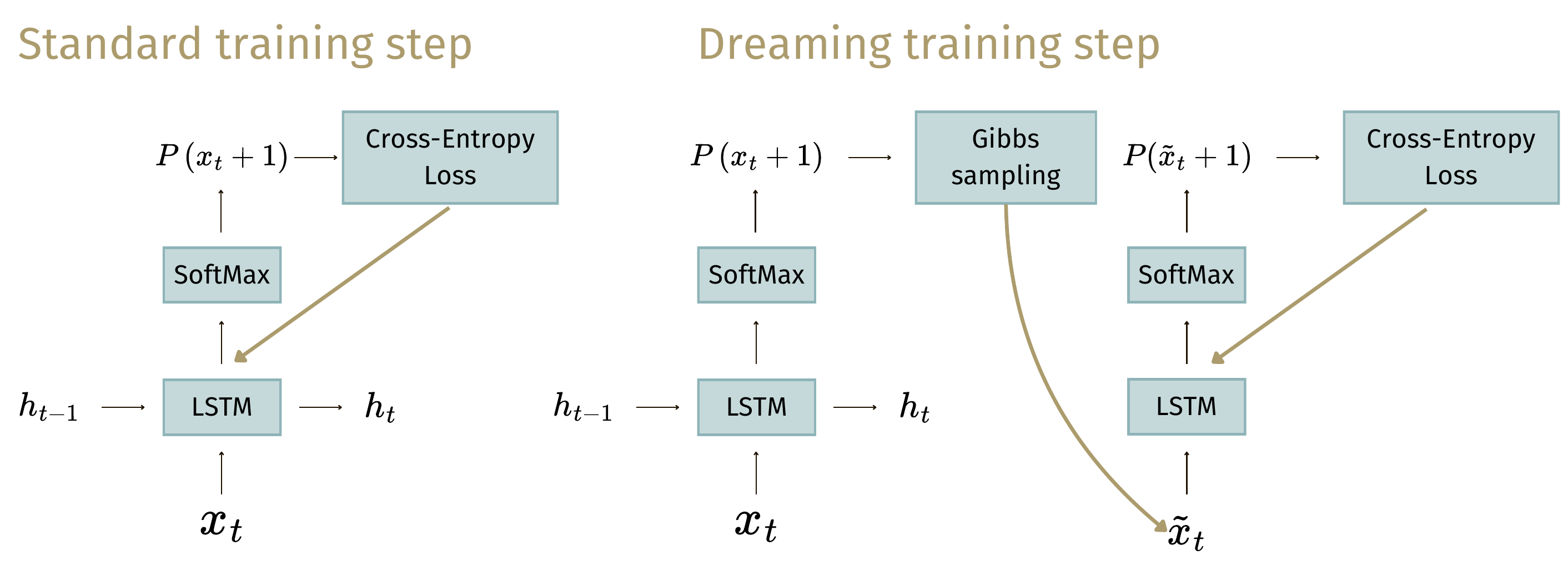}
    \caption{Diagram of the Dreaming Learning algorithm. The left side illustrates the standard training on the dataset, while the right side represents the Dreaming training step, where synthetically sampled data enhances general training for non-stationary data.}
      \label{fig:dreaming_maps}
\end{figure}

    Many natural systems, such as biological, social or technological systems, expand in size over time by continuously adding new elements to the existing ones. 
    As Stuart Kauffman~\citep{Kauffman2000-KAUI-4} notes, novelties arise when new recombinations of existing information are produced according to the system’s intrinsic evolutionary laws, thus augmenting the current set of elements. These new elements often represent slight variations of the objects that generated them, adding layers of complexity to the system. The potential for developing new objects and their proximity to some source information elements defines a concept known as the \textit{Adjacent Possible}. This notion describes a constantly evolving set of possibilities the system can explore and integrate. 
    The complexity inherent in such systems not only drives a gradual expansion of elements but also allows newly added components to significantly influence the system’s future evolution, making these systems open-ended and non-stationary as they continually adapt to internal and external stimuli.  However, incorporating new data into machine learning systems poses key challenges, especially regarding issues like catastrophic forgetting and continual learning~\citep{mccloskey1989catastrophic, ratcliff1990connectionist, hadsell2020embracing, 10444954}. 
    The primary challenge lies in preserving the memory of previously learned data without being disrupted or overwritten by introducing new datasets. This issue becomes even more challenging when the latest data contains information with statistical properties that differ from the old paradigm. In such cases, it is vital for the system to rapidly adapt its parameters to accurately represent the new paradigm while ensuring that previously acquired information is retained.
    Achieving this balance requires sophisticated mechanisms that enable the system to learn and adapt continually while minimizing the risk of forgetting essential information from earlier learning phases.
    \newline
    In line with this theoretical framework, we propose an innovative algorithm to explore the Adjacent Possible configurations of a neural network during the sequential learning of symbolic time series. This algorithm is aimed to leverage artificial temporal sequences generated through the probabilistic model the machine learned up to a given point in time to be better prepared for valuable novelties. These synthetic sequences are generated through Gibbs sampling, modulated by a carefully selected temperature parameter, allowing for a controlled exploration of the space of possibilities~\citep{murphy2013machine, wang2019bert}. The method seeks to computationally realize the theoretical concept of the Adjacent Possible by exploring sequences compatible with the existing knowledge acquired by the neural system. An intriguing aspect of this approach is that during the exploration phase, the system temporarily detaches from the external input data while generating artificial to be subsequently learned by the system. This unique feature inspired the name "Dreaming Learning" through an analogy with the dream sequences produced by the brain during REM sleep phases~\citep{HOEL2021100244}. By mimicking dreaming processes, our approach offers a novel mechanism for neural networks to restructure and enhance their understanding internally, enabling more flexible and robust adaptation to new learning challenges. This dreaming-like process facilitates the consolidation of existing knowledge and paves the way for creative exploration and innovation within artificial neural systems. The algorithm is schematized in \autoref{fig:dreaming_maps}.
\section{The Dreaming Learning Algorithm}\label{sec:The Dreaming Learning Algorithm}
    Dreaming Learning is a training method where the network generates some of its training data and learns from them in addition to the standard training dataset. Two conditions are essential for this method: the network's output must be transformable into its input dimensionally, and the network must be probabilistic. 
    \newline
The process involves two phases: first, the network is trained in the usual way by predicting the next element in a sequence and adjusting based on the cross-entropy loss. Then, in the Dreaming Learning step, the network generates a new synthetic sequence sampled from the current output distributions. After the generation, the network is trained again using the synthetically generated sequences.
\newline
We can state that, as shown in the \autoref{appendix:Bayes}, in the case of the Dreaming Learning phase, we have:
\begin{equation} \label{eq:1}
    \log P(\Omega| x_{t+1}, \bold x_t) \approx \log P(x_{t+1}|\bold x_t, \Omega) + \log P(\tilde{\bold x}_t| \Omega) + \log P(\Omega),
\end{equation}
where $\Omega$ are the weights, $\bold{x}_t$ are the data, and  $\tilde{\bold x}_t$ are the network-generated data. In contrast to classical Bayesian learning, the substitution with \(\tilde{\bold{x}}_t\) in \(P(\tilde{\bold{x}}_t|\Omega)\), now depending on \(\Omega\), makes this term participate in the general maximization of \autoref{eq:1}. We use Gibbs sampling at each iteration to generate the Dreaming sequence, which is conditioned by a specific sampling temperature \(T\), changing the output distribution in the Softmax output layer of the network. The entropy of the output distribution increases in the case of \(T > 1\) and decreases otherwise, leading towards the uniform distribution for \(T \gg 1\). For step \( N \), the network generates sequences \(\tilde{x}_N^k = [\tilde{x}_N^k, \tilde{x}_N^{k-1}, \ldots, \tilde{x}_N^1]\) based on the statistical model built up to that step. The expectation is that for sufficiently large \( N \), \( P(\tilde{x}_N^k) \rightarrow P(x_k) \). For more theoretical background, please refer to~\autoref{appendix:Bayes}.
\newline
Although Dreaming Learning can be included in the data augmentation class, comparing it with other methods in the same category is challenging. Standard data augmentation methods \citep{MUMUNI2022100258} concern datasets whose elements are not characterized by non-stationary dynamics—such as images or general language models—which makes such a comparison inadequate.
\section{Experiments}
We conduct two types of experiments to evaluate the advantages of Dreaming Learning in terms of accuracy in learning non-stationary time series. The first assesses the behavior concerning regime changes in Markov chains, while the second evaluates real-world symbolic series with local changes in stationarity, such as in language models. 
\subsection{Markov Chains' regime shift}
\begin{figure}[h]
    \centering
    \begin{minipage}[b]{0.48\textwidth}
        \centering
        \includegraphics[width=\textwidth]{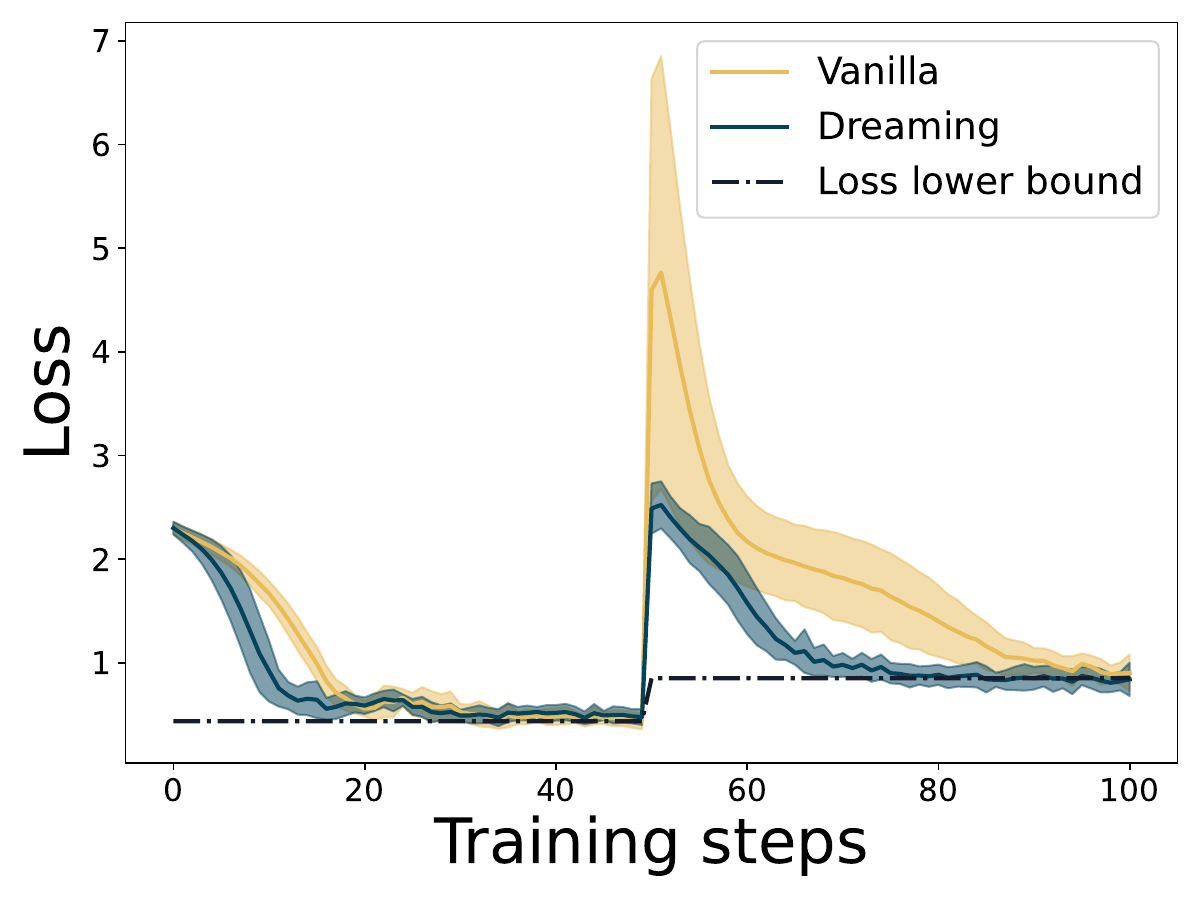}
        \caption{Example of training averaged on 20 simulations for \(T_s=1.5\) and initial and final normalized entropies of 0.3 and 0.6, respectively. The central peak in the loss corresponds to the regime shift for the Markov chain. The shaded areas are the losses' standard deviations.}
        \label{fig:meanlosses}
    \end{minipage}
    \hfill
    \begin{minipage}[b]{0.48\textwidth}
        \centering
        \includegraphics[width=\textwidth]{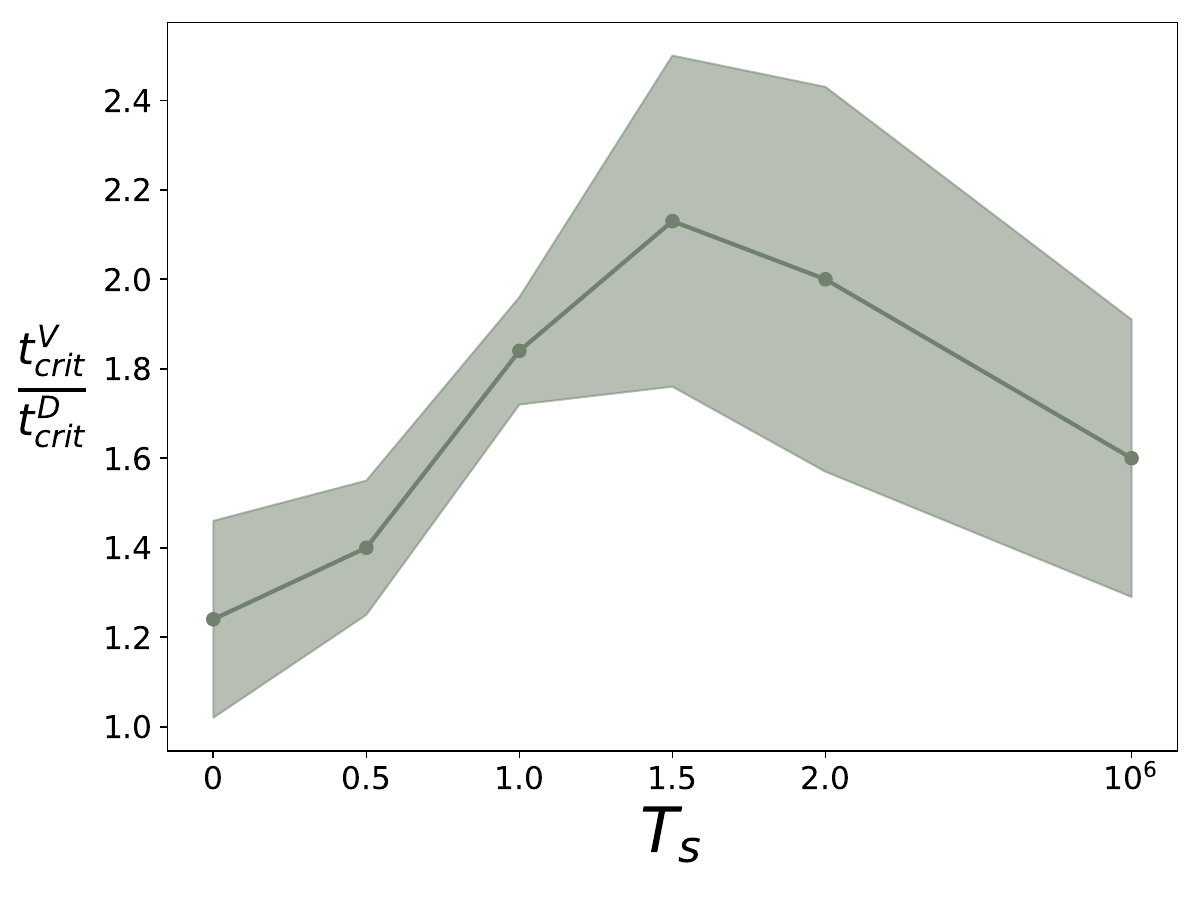}
        \caption{Mean ratio between Vanilla and Dreaming critical time $t_{crit}$ for different initial and final limit entropies of the Markov transition matrices. Every point is averaged over 640 training simulations. The shaded area is the ratio's standard deviation.}
        \label{fig:ratiotemperature}
    \end{minipage}
\end{figure}

Markov chains represent discrete stochastic systems in which the probability of transitioning from one state to another depends solely on the current state, making the system memoryless. Formally, this property is expressed as  $
P(X_{n+1}=x \mid X_1=x_1, X_2=x_2, \ldots, X_n=x_n) = P(X_{n+1}=x \mid X_n=x_n).$ Introducing novelties into such a system involves altering the transition matrix, which leads to a sudden shift in the limit probability distribution, thereby requiring the network to adapt to the new conditions.
\newline
To assess the efficacy of the Dreaming Learning approach, we compare the regular (Vanilla) and the Dreaming training algorithms on the same neural architecture.
The Dreaming Learning step is applied as described in \autoref{sec:The Dreaming Learning Algorithm}. 
\autoref{fig:meanlosses} shows the mean training results across 20 simulations of Vanilla and Dreaming networks. A sudden change in the Markov transition matrix shifts the lower bound of the limit distribution entropy from 0.3 to 0.6, altering the system's state dynamics and causing a spike in the loss. However, Dreaming Learning outperforms the Vanilla approach in convergence speed and overall loss after the spike, adjusting first to the new conditions.
Moreover, since different sampling temperatures $T_s$ can produce different outcomes depending on the Markov chain's limit distribution entropy, we compare the \(t_{crit}\), which indicates how quickly the loss function approaches its lower bound, of the exploring Dreaming Learning network with the Vanilla network across a range of initial and final entropy values. This helps us identify scenarios where Dreaming Learning outperforms the Vanilla approach, i.e., where this ratio is \(>1\).
We repeat this evaluation for different sampling temperatures to determine optimal temperature ranges. The Markov chains used in the training were uniformly selected based on the entropies of their limit distributions, with normalized entropy values ranging from 0.1 to 0.8 and an entropy step of 0.1. For each pair of initial and final entropy values, we conducted 10 simulations, resulting in a total of 640 training runs, both vanilla and dreaming, for each point of \autoref{fig:ratiotemperature}. The performance is quantitatively assessed by averaging the entropy values and analyzing the dependence on the sampling temperature. As we can see in \autoref{fig:ratiotemperature}, Dreaming Learning shows more than a twofold improvement over the Vanilla network around the sampling temperature \(T_s=1.5\). Notably, the advantage of Dreaming Learning begins to emerge at \(T_s = 0.5\) and diminishes at higher temperatures. This optimal region, related to the concept of the Adjacent Possible, represents the temperature range where exploration is most effective. In contrast, higher temperatures lead to a flatter sequence distribution, making the Dreaming Learning network behave more like the Vanilla network with added noise.
 
\subsection{Regime shift in language models}
To assess the effectiveness of Dreaming Learning in a real-world scenario with non-stationary data, we constructed a small word-level language model using an LSTM network \citep{HochSchm97}. We selected 16 public domain books, creating a single sequence of 3,128,772 words (tokens) with a total vocabulary of 43,138 tokens. The sequential transition from one book to the next characterizes the non-stationary feature in this experiment. In general, we expect the neural network to be able to learn the general statistics of the English language, such as grammatical and syntactic rules, while also focusing on changes in style, characters, and setting characterizing every novel. This requires the network to predict and anticipate new word sequences, even when they differ from previous ones, and may include unexpected tokens. In this context, the transition from one book to the next produces a non-stationary process represented by a regime shift of the language's statistical properties, clearly visible by the spikes in the loss. The neural network architecture is a four-layer LSTM with 400 cells per layer and two linear layers for input and output. The network hyper-parameters have been selected after testing different architectures and assessing the test loss on one book not included in the training and validation sets.
\begin{figure}[htb]
    \centering
    \begin{subfigure}[b]{\textwidth}
        \centering
        \includegraphics[width=0.5\linewidth]{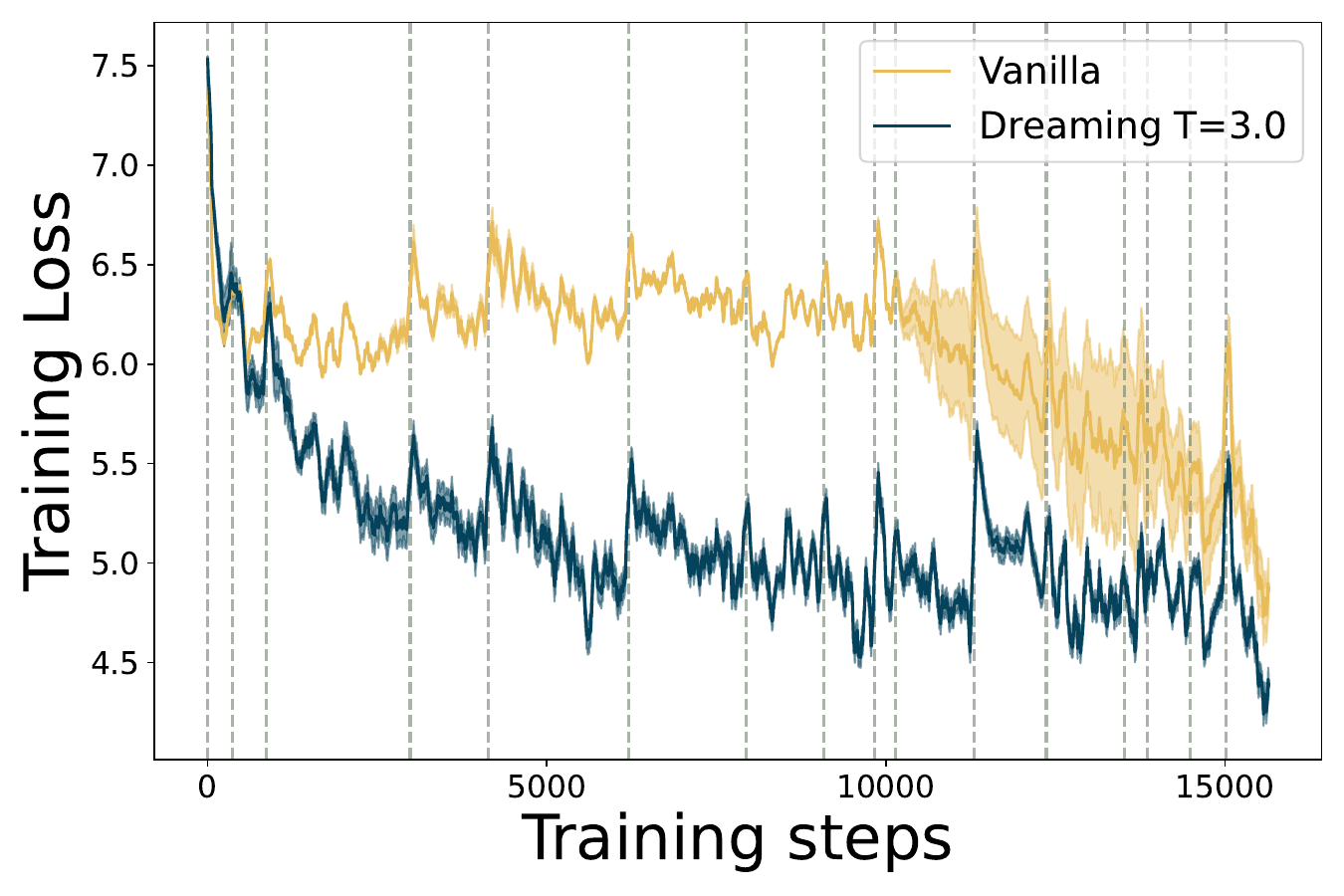}%
        \hfill
        \includegraphics[width=0.5\linewidth]{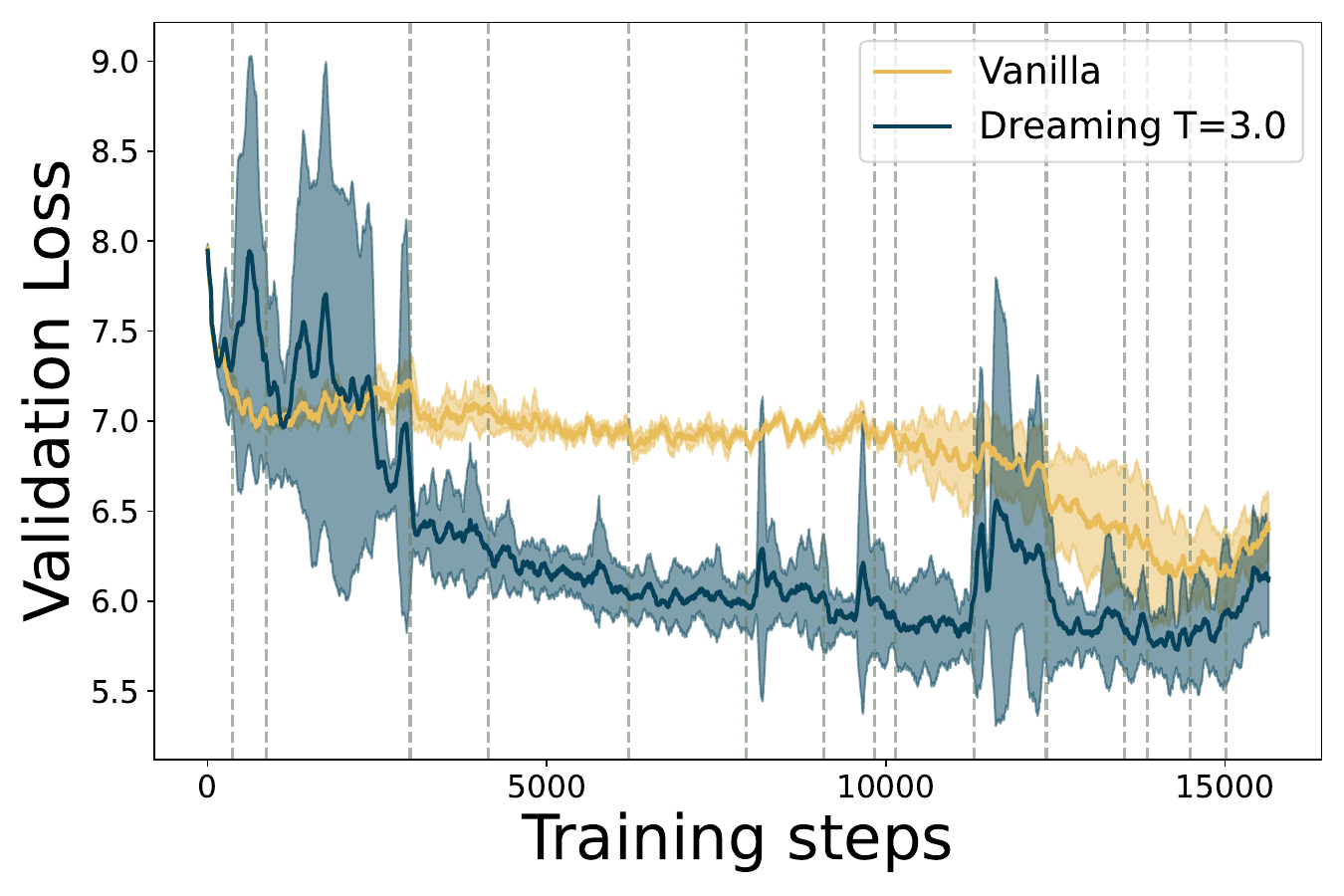}
        
\end{subfigure}  
\caption{Comparison between Vanilla and Dreaming-powered neural network (left) training and (right) validation loss (sampling temperature = 3) averaged on 10 independent trainings. The vertical dashed lines indicate the beginning of the training books.}
\label{fig:text_losses}
\end{figure}
As shown in \autoref{fig:text_losses} (left panel), the Dreaming network rapidly achieves significantly lower training loss values than the Vanilla network. The quality of learning is confirmed in \autoref{fig:text_losses} (right panel) by a lower validation loss value carried out on three reference books not used in the learning dataset. 
\newline
At the end of the standard training on the 16 books comprising the original training dataset, the Vanilla and Dreaming networks generated a text sequence of approximately 3 million words. To assess the goodness of these generated sequences, we evaluated the degree of correlation in the artificial text using Hurst's exponent \citep{Hurst1951LongTermSC}, in analogy with the inspection techniques shown on \citet{lippi2019natural}. Hurst's exponent measures long-term memory of time series through auto-correlations and the rate at which these decrease as the lag between pairs of values increases. While Hurst's exponent is 0.710 for the actual textual sequence, the Vanilla-generated text shows a value equal to 0.601, and, in contrast, the Dreaming network text has a Hurst's exponent equal to 0.662, with a sensitive increase of the generated text correlation equal to 29\%. Finally, the Dreaming Learning contribution to the generation of more natural language is given by the assessment of the Heap's exponent~\citep{tria2014dynamics}, related to the rate of innovation of a symbolic sequence. This exponent is 0.54 for real text, while for the Dreaming generated sequence, it is equal to 0.549, compared to the value of 0.453 from the Vanilla network.

\section{Conclusions}
Dreaming Learning incorporates an exploration phase into regular neural network training. This technique enhances the handling of non-stationary symbolic time series, where the system must process new elements or sequences. A key aspect of Dreaming Learning is the network's ability to adapt to new, statistically distinct information related to previously encountered data, aligning with the Adjacent Possible theory. This approach improves network adaptability to new sequences. It optimizes training and validation losses, providing a valuable framework for training on open-ended and evolving sources. This leads to an effective exploit-explore process, where the Adjacent Possible is explored in the Dreaming phase, allowing better training and faster inclusion of new regimes.

\newpage
\section*{Acknowledgments}
We would like to thank PNRR MUR project PE0000013-FAIR for their support.
\bibliography{biblio}
\bibliographystyle{plainnat}
\newpage
\appendix
\section{Probabilistic description of Dreaming Learning}\label{appendix:Bayes}
Given a dataset \(D=\{\bold y, \bold x\}\), where \(\bold y, \bold x \in \mathbb{R}^d\) and \(|D|=N\) is the amount of available data, training a machine learning model is equivalent to find the best system parameters configuration \(\Omega\) to maximize the quantity:
\begin{equation}
    P(\bold y|\bold x, \Omega)
\end{equation}

\noindent For old approaches, like the maximum likelihood method, such maximization was equivalent to maximizing the term \(P(\Omega|\bold y, \bold x)\):
\begin{equation} \label{likelihood}
    \max_{\Omega}[P(\Omega|\bold y, \bold x)] = \max_{\Omega}[P(\bold y|\bold x, \Omega)]
\end{equation}

\noindent This mathematical relationship is valid when a uniform distribution is sufficient to describe the priors related to the dataset, a situation that, in general, is hardly verified. Bayes' Theorem gives a more general description:
\begin{equation} \label{bayes1}
    P(\Omega|\bold y, \bold x) = \frac{P(\bold y,\bold x| \Omega) \cdot P(\Omega)}{P(\bold y, \bold x)} = \frac{P(\bold y|\bold x, \Omega) \cdot P(\bold x,|\Omega)\cdot P(\Omega)}{P(\bold y, \bold x)}
\end{equation}
Since $P(\bold x,|\Omega)\cdot P(\Omega) = P(\bold x,\Omega)$:
\begin{equation} \label{bayes2}
    P(\Omega|\bold y, \bold x) = \frac{P(\bold y|\bold x, \Omega) \cdot P(\bold x,\Omega)}{P(\bold y, \bold x)}  
\end{equation}
\noindent The maximization of \ref{bayes2} can get rid of the denominator since it does not depend on \(\Omega\), leading to the following equality:
\begin{equation}
    \max_{\Omega}[P(\Omega|\bold y, \bold x)] = \max_{\Omega}[P(\bold y|\bold x, \Omega) \cdot P(\bold x, \Omega)]
\end{equation}

\noindent whose logarithmic description is:
\begin{equation}
    \log P(\Omega|\bold y, \bold x) \approx \log P(\bold y|\bold x, \Omega) + \log P(\bold x, \Omega)
\end{equation}

\begin{equation}
    \log P(\Omega|\bold y, \bold x) \approx \log P(\bold y|\bold x, \Omega) + \log P(\bold x| \Omega) + \log P(\Omega)
\end{equation}

\noindent This relationship is much more suitable than \ref{likelihood} since it considers the specific shape of data distribution, which allows finding better-optimized solutions. The disadvantage of this approach is that the knowledge of such prior is commonly missing, and it can only be estimated by original data and its intersection with the model parameters \(\Omega\).

In the case of a time series, \(\bold y \Rightarrow x_{t+1}, \bold x \Rightarrow [x_{t}, x_{t-1}, \ldots, x_1]\), that is the system output is a single value element given the history of the time series described by the sequence leading up to it. In our case, \(x_t\) is a discrete value variable from a given vocabulary. For example, the time series may represent characters or words from a corpus of texts or the discretized version of a real variable according to a related quantization rule.

For a time series, the Bayesian training description is:
\begin{equation}
    \log P(\Omega| x_{t+1}, \bold x_t) \approx \max_{\Omega} [ \log P(x_{t+1}|\bold x_t, \Omega) + \log P(\bold x_t, \Omega)
\end{equation}

\begin{equation} \label{eq:8}
    \log P(\Omega| x_{t+1}, \bold x_t) \approx \log P(x_{t+1}|\bold x_t, \Omega) + \log P(\bold x_t| \Omega) + \log P(\Omega)
\end{equation}

where, for simplicity of notation, \( \bold x_t = [x_t, x_{t-1}, \ldots, x_1] \).

In the classic learning approach, \( P(\bold x_t | \Omega) = P(\bold x_t) \) since the training data does not depend on the network parameters \(\Omega\). However, the conditioning term is held in the case \(\bold x_t\) is generated by the network, contributing to the Bayesian probability maximization. Now on, the generated data is described by \(\tilde{\bold x}_t\).

Equation \ref{eq:8} can be rewritten as

\begin{equation} \label{eq:9}
    \log P(\Omega| x_{t+1}, \bold x_t) \approx \log P(x_{t+1}|\bold x_t, \Omega) + \log P(\tilde{\bold x}_t| \Omega) + \log P(\Omega)
\end{equation}

where \(\tilde{\bold x}_t \Rightarrow [\tilde{x}_{t}, \tilde{x}_{t-1}, \ldots, \tilde{x}_1]\).

We use Gibbs sampling at each iteration in Dreaming Learning to generate the Dreaming sequence. Such a generation is conditioned by a specific sampling temperature \(T\), changing the output distribution in the Softmax output layer of the network. The entropy of the output distribution increases in the case of \(T > 1\) and decreases otherwise, leading towards the uniform distribution for \(T\gg 1\). 

Generating sequences to be plugged in equation \ref{eq:9} allows the term \(\log P(\tilde{\bold x}_t|\Omega)\) to have the role of a regularizer related to the data prior, which is absent in the classic training algorithms. The first term of the right-hand equation is Bayes' likelihood, which is related to conventional training on the reference data set. The second term has a twofold interpretation. On one side, it represents the same statistical law of the first term for the previous sequences at time \(t\) because of the chain rule \( \log P( \bold x_t, \Omega )  = \sum_{k=0}^{k=t-2} \log P(x_{t-k} | \bold x_{t-k-1}, \Omega) \). Therefore, such a term is evaluated during the traditional training, as is the likelihood term. On the other hand, to be properly assessed, the prior assessment would require the evaluation of all the possible sequences, which is not feasible because of the limited availability of sequences in the data set available for learning. Moreover, using a sampling temperature \(T>1\) allows the system to explore new sequences whose compatibility with the data statistical properties is driven by the temperature itself. Aim of such an approach is maximizing the posterior \(\log P(\Omega|x_{t+1}, \bold(x)_t\) by selecting the suitable temperature \(T\) for balancing the effects of the likelihood  \(\log P(x_{t+1}|\bold x_t, \Omega)\) and the data prior \(\log P(\tilde{\bold x}_t| \Omega)\).

At training step \(N\), the network can generate the sequence \(\tilde{\bold x}_k^N = [\tilde{x}_k^N, \tilde{x}_{k-1}^N, \ldots, \tilde{x}_1^N]\) according to the statistical model built up to that step. The likelihood term steers the neural network toward a model compatible with the learning data by trying to minimize the error between actual and expected outputs. At the same time, the prior term allows the network to test many possible sequences, synthetically increasing the data available to the neural network. The expectation is that for sufficiently large \(N\):

\begin{equation}
    \tilde{\bold x}_k^N \xrightarrow[N \to \infty]{} \bold x_k^N
\end{equation}

According to the above, we introduced a method to progressively estimate \(P(\bold x_t| \Omega)\) by leveraging the ability of probabilistic machine learning models to explore the distribution estimated during the training phase through recursive output data sampling.

\section{Cross-Entropy Loss analysis}
In a probabilistic neural network, the loss function used is the Cross-Entropy Loss between the expected probability distribution and the probability distribution predicted by the network, denoted as \(P_{\text{pred}}\). In the context of Dreaming Learning, the expected probability distribution is derived from the network's own generated data, denoted as \(Q_{\text{gen}}\). The loss function can be expressed as:

\begin{equation}
H(P_{\text{pred}}, Q_{\text{gen}}) = -\sum_{\tilde{\mathbf{x}}_k^N \in \tilde{\mathbf{X}}} P_{\text{pred}}(\tilde{\mathbf{x}}_k^N) \log P_{\text{pred}}(\tilde{\mathbf{x}}_k^N) + \sum_{\tilde{\mathbf{x}}_k^N \in \tilde{\mathbf{X}}} P_{\text{pred}}(\tilde{\mathbf{x}}_k^N) \log\left(\frac{P_{\text{pred}}(\tilde{\mathbf{x}}_k^N)}{Q_{\text{gen}}(\tilde{\mathbf{x}}_k^N, T)}\right)
\end{equation}
The first term is the Shannon entropy of the network's probability distribution \(P_{pred}\), the second term is the Kullback-Leibler (KL) divergence between \(P_{pred}\) and \(Q_{gen}\). \(T\) is the sampling temperature. The behavior of the loss function varies with the value of \(T\). When \(T = 1\), the right-hand term's role is primarily to increase the sample size used for training. If the sample size \(N\) and the sequence length \(k\) are sufficiently large, the contribution of this term becomes negligible, and the loss function essentially reduces to the entropy of the predicted distribution \(P_{\text{pred}}\). When \(T > 1\), the distribution \(Q_{\text{gen}}\) becomes flatter than \(P_{\text{pred}}\). The KL divergence term becomes significant, encouraging the network to adjust its weights to flatten \(P_{\text{pred}}\), thereby increasing the probabilities of rarer events relative to more common ones. When \(T < 1\), the distribution \(Q_{\text{gen}}\) becomes sharper compared to \(P_{\text{pred}}\). The learning process forces \(P_{\text{pred}}\) to focus more on the most probable events, potentially causing less common events to disappear from the predicted distribution. In summary, the temperature parameter \(T\) controls the balance between common and rare events in the predicted distribution. A higher \(T\) promotes exploration by assigning more weight to rare events, while a lower \(T\) emphasizes exploitation by focusing on the most likely events.

\end{document}